\documentclass[11pt]{futurenetworkandmobilesummit12}
\usepackage{amsmath}
\usepackage{amsfonts}
\usepackage{algorithm} 
\usepackage{algorithmic} 
\usepackage{fancyhdr}
\usepackage{lastpage}
\usepackage[pdftex]{graphicx}
\usepackage{graphicx}
\usepackage{subfigure}
\usepackage{setspace}
\usepackage{epsf}
\newcommand{\comments}[1]{} 

\title{ARIANNA: pAth Recognition for \\ Indoor Assisted NavigatioN \\ with Augmented perception}

\author{ Pierluigi GALLO$^1$, Ilenia TINNIRELLO$^1$, Laura GIARR\'E$^1$,  \\
Domenico GARLISI$^1$, Daniele CROCE$^1$, and Adriano FAGIOLINI$^1$ \\
$^1$\emph{DEIM, viale delle Scienze building 9, Universit\'a di Palermo,  90128 Palermo - Italy\\
Email: {\tt <name>.<SURNAME>@unipa.it} }
}

\begin{document}
\maketitle

\begin{abstract}
ARIANNA stands for pAth Recognition for Indoor Assisted Navigation with Augmented perception. It is a flexible and low cost navigation system for visually impaired people. Arianna permits to navigate colored paths painted or sticked on the floor revealing their directions through vibrational feedback on commercial smartphones.

\end{abstract}


\section{Introduction}
Arianna is the Italian name for Ariadne, Minos' daughter in Greek mythology. Her idea to help Theseus in overcoming the Minotauro and come out from the labyrinth is the basic inspiration for our work. Blind people are forced to live in a labyrinth of darkness and the only way to come out is to create a visual map of surroundings. Depending on the severity of the impairment, recreation of such map can be more difficult having (i) partially sighted people, (ii) low vision, that are unable to read a newspaper at a normal distance even with corrective eyeglasses or contacts; (iii) legally blind whose vision is less than a defined threshold, (iv) totally blind people that have no vision at all. Many factors can be the cause of vision impairment, from accidents, diabetes, retinitis and lack of A vitamin. Severe impairments can be very impacting on the quality of life of such people because daily tasks are made more difficult or even impossible. 

Visually impaired people refine remaining senses to perceive the environment making a strong use of hearing and touch senses to compensate their lack of sight. Navigation towards a destination can be realized by sensing the immediate surroundings for impediments to travel (e.g., obstacles and hazards) but also creating a map that goes beyond it. Navigation methods can be classified accordingly to the quantity used by the traveler’s brain: position, speed and acceleration \cite{}. The contribution provided by this work regards the definition of a system for the autonomous navigation of blind people in unfamiliar environments.

\section{System requirements}
Assistive tools for visually impaired have specific requirements in terms of reaction time (they require to run in real-time to be useful) so they need an adequate refresh frequency. Tools must be light-weight, portable, low-power and low-cost and should require minimum training time. Solutions based on off-the-shelf devices can be easily spread, even better if the used devices are already available to people. 
The availability of the system should be both in outdoor and in indoor, where the unavailability of GPS-based solutions introduces an extra challenge. Most of current assistive tools employ vocal indications to inform the traveler about position, environment, displayed information. However, visually impaired use hearing to catch information on the near environment,  so audio instructions from a piloting system, especially if continuous and repetitive are perceived as a distraction and an overload both by visually impaired and by other people in the neighborhood. It results in avoiding audio indications in favor of the remaining alternatives, among which tactile is the most prominent. 

\section{Navigation system description}
The ARIANNA navigation systems is composed by four main components: (i) ambient instrumentation; (ii) sensors; (iii) data transport network; (iv) path server; (v) user interface.
The ambient instrumentation is quite simple and low cost: colored tapes can be easily sticked on the floor or carpets define to define different paths. This is the only dedicated instrumentation applied to the ambient, because the WiFi network has not dedicated requirements.
The only sensor used in the ARIANNA system is the camera. Most of common smartphones on the marketplace are equipped with a camera; it is used to reveal the presence of lanes on the floor and acts as a visual to haptic transducer.
The data transport network does not require specific adaptations but is a facility that permits communication between the phone and the ARIANNA server. The server is used to provide localization information, correlation between paths and points of interest, routing towards destination. The presence of the server and the wireless network is necessary only in case the application is unaware about the building topology and its deployed paths. On the contrary, if the application loaded on the phone has such information locally available, the presence of network and server is optional (even if flexibility is possible only with those elements, as explained later on).  
The user interface employes tactile stimuli, as better described in the following. As reported in Figure \ref{f:scheme}, the ambient is instrumented with colored lines on the floor; QRcodes are settled close to points of interest and on line intersections. They provide information on the right line to follow in order to get to the desired destination.

\begin{figure}[t]
\centering
{\includegraphics[width=0.8\textwidth]{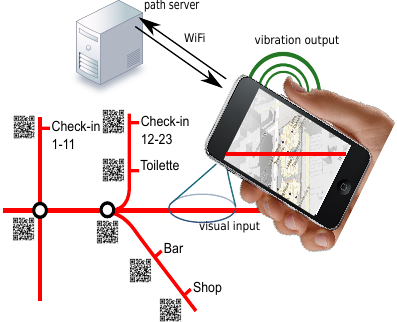}}
\caption{ARIANNA navigation system description}
\label{f:scheme}
\vspace{-0.4cm}
\end{figure}

\subsection{Tactile Interface}
The tactile interface is a key point of the system. The behavior of the haptic feedback can be summarized as follows: the camera continuously grab the scene in front of the person. The user touches the screen of his smartphone: if the touched point falls into the area of the screen where the path is reproduced, than the whole phone vibrates. Vibration is a native functionality of the phone. Unlike other approaches in haptic interfaces, our solution does not need a selective vibration of the touched point (that is also difficult to obtain and requires special piezo-electric materials, etc.). The user perceives vibration only when touching 'sensitive areas' of the screen, so it associates the vibration to them, as if only those points vibrates.

\subsection{Path encoding}
Paths can intersect each other forming a planar graph where intersections are nodes of the graph. Any path segment (the graph edges) may be deployed with two parallel strips with different colors, so the ordered couples (color1, color2) and (color2, color1) encode both direction and orientation. Using bar codes it is possible to encode relevant information regarding the edges (as for example the distance from/to the extremes of the segment).

\subsection{The path server}
 The path server stores and retrieves information from a path repository via the url printed into the QRcode. The content pointed out by the (fixed) url can be changed on the fly with a simple update on the server. Such flexibility permits path adaptation required by topological changes due to maintenance or load balancing. 
When the smartphone detects a QRCode on the path, it immediately run an http request to the server using the url inside the QRcode. The server knows the position of the user (because of its proximity to the QRCode position) and sends back to the smartphone the next edge to follow. In facts, among all paths deployed in the building, thanks to the indications provided by the path server, the smartphones provides haptic feedback only towards the 'enabled' paths according to the server indication.

\section{Usage description}

\begin{figure}[h]
\centering
\subfigure[]{\includegraphics[width=0.3\textwidth]{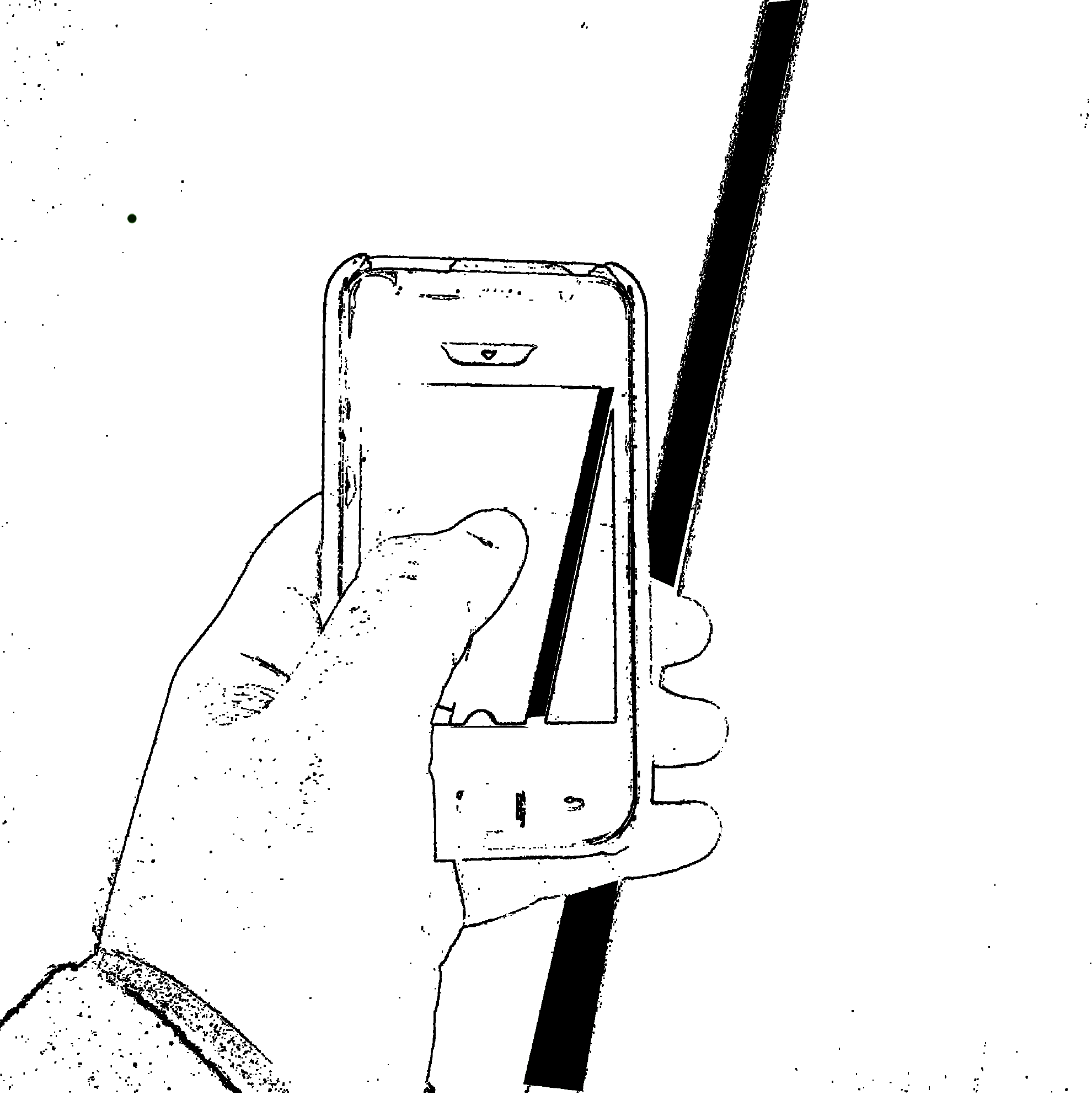}} 
\subfigure[]{\includegraphics[width=0.3\textwidth]{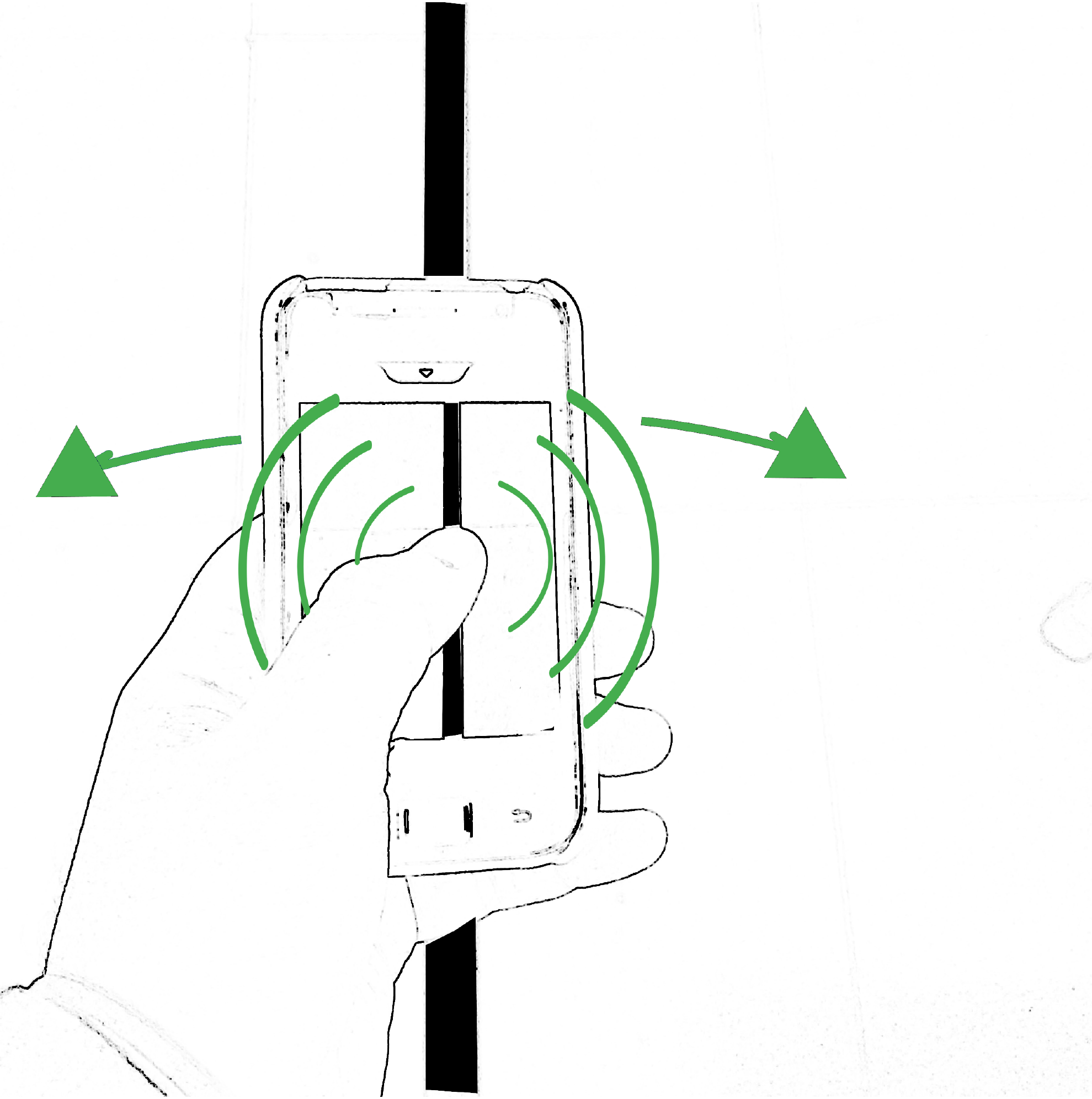}} 
\subfigure[]{\includegraphics[width=0.3\textwidth]{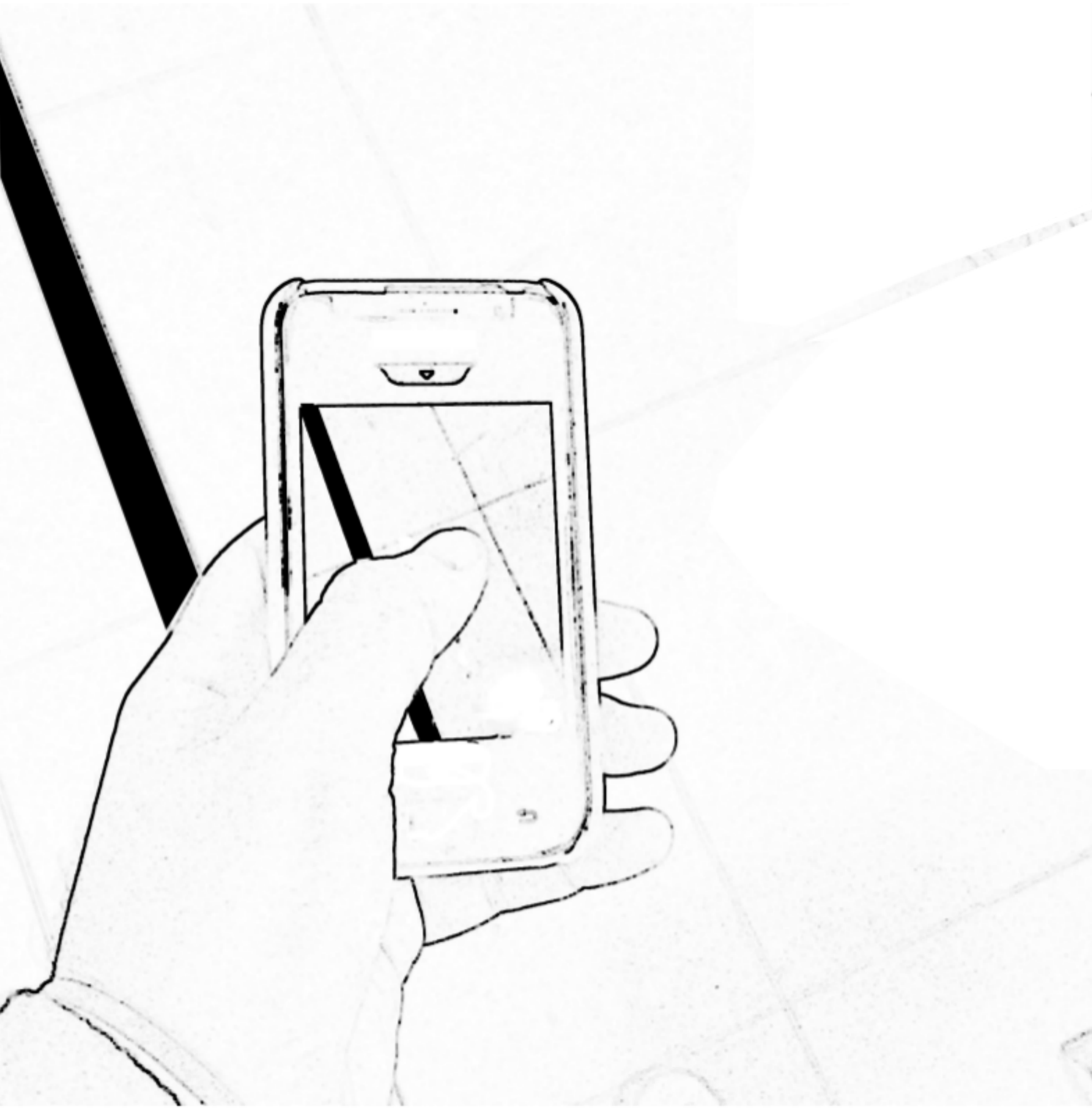}} 
\caption[]{Scanning the environment with fixed finger as using a cane from left side of the lane (a), crossing the line (b), right side of the lane.}
\label{f:scanning}
\vspace{-0.4cm}
\end{figure}

Because of its easiness to deploy, ARIANNA can be implemented in whatever indoor and outdoor scenarios, as for example airports, schools, hospitals, museums, parks, and sidewalks. Furthermore, ARIANNA is easy to use because it employes the smartphone as a visual-to-vibrational translator, transforming visual lines painted on the floor in vibrational information. As shown in figure \ref{f:scanning}, the visually impaired person can walk normally scanning her near surroundings with the phone with a left-to-right and right-to-left movement like the traditional cane movement. In green it is reported the vibrational effect when crossing the line: the position of the hand relative to the body reveals, thanks to the proprioception, what is the direction to follow. Change in the direction of the lane are revealed by the walking person, which naturally orientate her body in the right direction. Vibrational stimuli are provided only when the finger touches the lane, so the screen can be also scanned moving the finger from left to right as following a written page.

\vspace{-0.4cm}

The system has been shown during the workshop organized by the Andrea Bocelli Foundation in Boston \cite{abf2013}.

\begin{figure}[t]
\centering
{\includegraphics[width=0.8\textwidth]{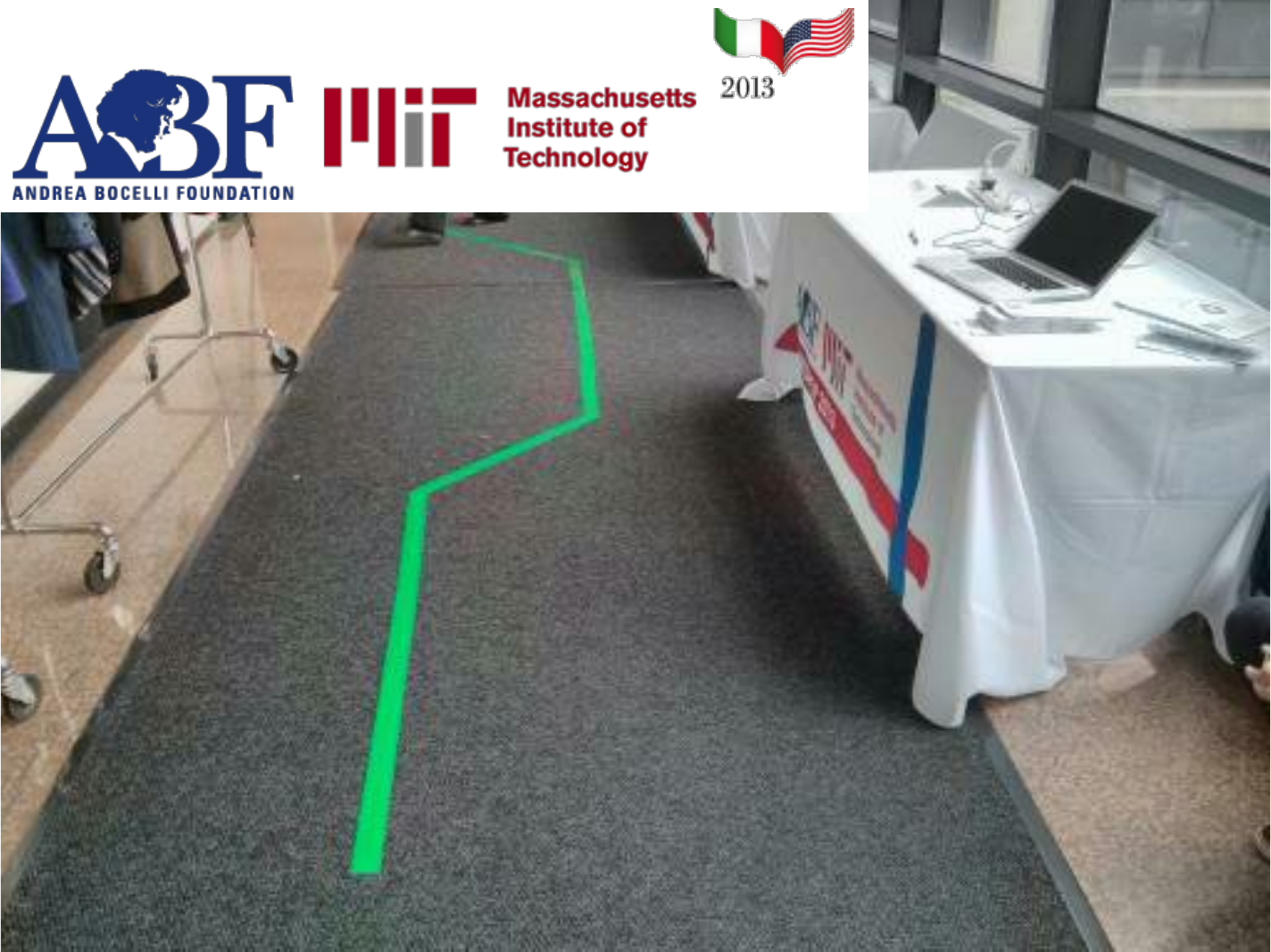}}
\caption{The simple test path provided provided during the ABF workshop}
\label{f:scheme}
\vspace{-0.4cm}
\end{figure}

\section{Related work}
Several assistance tools exists to overcome visual impairments, most of them exploiting vocal instructions to inform the traveller about his position and the near environment. For example, virtual acoustic displays and verbal commands issued by a synthetic speech display are used in \cite{navi}. AudioGPS \cite{audiogps} and Melodious Walkabout\cite{melod} use audio cues to provide information on the surrounding environment. Dead-reckoning techniques are employed in Navatar\cite{Navatar} where users interact with the application and help correcting possible navigation errors. In \cite{visual}, vibrational feed-back is given by a special glove in the Finger-Braille language. This system requires some dedicated hardware and is specific to the language used. RF-PATH-ID \cite{rfpath}, instead, is based on disseminating passive RFID tags and using a dedicated reader to acquire information on the user location. More examples and detailed information on indoor localization techniques may be found in \cite{Fallah06022013}.

Haptic principles and a list of possible applications are presented in  \cite{haptic}. Because of the wide use of haptic interfaces, recently some benchmark metrics have been proposed, based on a combination of physical and psychophysical data  \cite{system}. In some recent works, haptics have been studied regarding frictional forces arising from the stroke of a finger moving on a surface  \cite{loomis,mobic}. Other touch interfaces use tangential skin displacement at the fingertip (stimulus speed and displacement length) in order to communicate direction or displaying static friction in haptic applications  \cite{solazzi}. Amplitude-modulated vibrotactile stimuli are commonly used; in  \cite{park} it is described an analysis able to map amplitude-modulated vibrations on their perceptual relations. In that work it is shown that the perception of vibration increases at very low modulation frequencies (1–10 Hz), while decreases for higher modulation frequencies (10–80 Hz). Amplitude-modulated signals can be discriminated by their envelope waveform instead of their spectral energy distributions. Multiple vibrators are used in  \cite{heuten}, where the traveller is guided by a belt with several vibrators; among which the one vibrating indicates the suggested direction to follow. PocketNavigator, introduced in  \cite{pocketnav}, is an Android application that gives a continuous feedback on direction and distance by encoding such quantities in vibration patterns.

\section{Future work}
As a future work we plan to introduce infra-red paths, which have no esthetic impact on the environment. Cameras on board smartphones are also sensible to infra-red bands, such a feature is currently underused. 

\section{Acknowledgment}
This work has been presented in the exhibition at the $2^{nd}$ workshop
challenges organized by the Andrea  Bocelli Foundation at MIT - Boston.

\bibliographystyle{ieeetr}

\end{document}